\documentclass[journal,twoside,web]{IEEEtran}
\usepackage[cmex10]{amsmath}
\usepackage[separate-uncertainty=true,multi-part-units=single]{siunitx}
\usepackage{tikz}
\usepackage{pgfplotstable}
\usepackage[draft]{changes}
\usepackage{graphicx}
\usepgfplotslibrary{groupplots}
\usepgfplotslibrary{colormaps}
\pgfplotsset{colormap/jet}
\usetikzlibrary{shapes.misc}
\tikzset{cross/.style={cross out, draw=black, minimum size=2*(#1-\pgflinewidth), inner sep=0pt, outer sep=0pt},cross/.default={1pt}}
\usepackage[acronym]{glossaries}
\usepackage{generic}
\usepackage{cite}
\usepackage{textcomp}
\usepackage{caption}
\makenoidxglossaries
\newacronym{dof}{DOF}{Degrees of Freedom}
\newacronym{rmc}{RMC}{Repeated Measures Correlation}
\newacronym{rmse}{RMSE}{Root Mean Square Error}
\newacronym{mee}{MEE}{Metabolic Energy Expenditure}
\newacronym{cmc}{CMC}{Coefficient of Multiple Correlation}
\newacronym{omc}{OMC}{Optical Motion Capture}
\newacronym{sa}{SA}{Sensitivity Analysis}
\newacronym{gsa}{GSA}{Global Sensitivity Analysis}
\newacronym{lsa}{LSA}{Local Sensitivity Analysis}
\newacronym{mc}{MC}{Monte Carlo}
\newacronym{oat}{OAT}{One-factor-At-a-Time}
\newacronym{lstm}{LSTM}{Long Short-Term Memory}
\newacronym{ann}{ANN}{Artificial Neural Network}
\def\BibTeX{{\rm B\kern-.05em{\sc i\kern-.025em b}\kern-.08em
    T\kern-.1667em\lower.7ex\hbox{E}\kern-.125emX}}
\usetikzlibrary{patterns}
\usepackage[separate-uncertainty = true,multi-part-units=single]{siunitx}
\usepackage[caption=false, font=footnotesize]{subfig}
\pgfplotsset{compat=1.17}
\definecolor{fau-fau-250}{rgb}{0.7490196078431373, 0.795278738946559, 0.8553171856978086}
\definecolor{fau-fau-375}{rgb}{0.6235294117647059, 0.6929181084198386, 0.7829757785467129}
\definecolor{fau-fau-625}{rgb}{0.37254901960784315, 0.48819684736639757, 0.6382929642445214}
\definecolor{fau-fau-dark}{rgb}{0.01568627450980392, 0.11764705882352941, 0.25882352941176473}
\definecolor{fau-fau-light}{rgb}{0.37254901960784315, 0.48627450980392156, 0.6392156862745098}
\definecolor{fau-fau_dark-125}{rgb}{0.876478277585544, 0.889273356401384, 0.9069896193771626}
\definecolor{fau-fau_dark-250}{rgb}{0.7529565551710881, 0.7785467128027682, 0.8139792387543252}
\definecolor{fau-fau_dark-375}{rgb}{0.629434832756632, 0.6678200692041523, 0.7209688581314879}
\definecolor{fau-fau_dark-625}{rgb}{0.3823913879277201, 0.4463667820069205, 0.5349480968858131}
\definecolor{fau-fau_dark}{rgb}{0.01568627450980392, 0.11764705882352941, 0.25882352941176473}
\definecolor{fau-fau_light-125}{rgb}{0.9212610534409842, 0.9355324875048059, 0.954725105728566}
\definecolor{fau-fau_light-250}{rgb}{0.8425221068819685, 0.8710649750096117, 0.9094502114571319}
\definecolor{fau-fau_light-375}{rgb}{0.7637831603229527, 0.8065974625144176, 0.8641753171856978}
\definecolor{fau-fau_light-625}{rgb}{0.6063052672049212, 0.6776624375240292, 0.7736255286428296}
\definecolor{fau-fau_light}{rgb}{0.37254901960784315, 0.48627450980392156, 0.6392156862745098}
\definecolor{fau-fau}{rgb}{0.0, 0.1843137254901961, 0.4235294117647059}
\definecolor{fau-med-125}{rgb}{0.8745098039215686, 0.954725105728566, 0.9847443291041907}
\definecolor{fau-med-250}{rgb}{0.7490196078431373, 0.9094502114571319, 0.9694886582083814}
\definecolor{fau-med-375}{rgb}{0.6235294117647059, 0.8641753171856978, 0.9542329873125721}
\definecolor{fau-med-625}{rgb}{0.37254901960784315, 0.7736255286428296, 0.9237216455209535}
\definecolor{fau-med-dark}{rgb}{0.0, 0.3803921568627451, 0.6274509803921569}
\definecolor{fau-med-light}{rgb}{0.37254901960784315, 0.7725490196078432, 0.9254901960784314}
\definecolor{fau-med_dark-125}{rgb}{0.8745098039215686, 0.922245290272972, 0.9532487504805844}
\definecolor{fau-med_dark-250}{rgb}{0.7490196078431373, 0.8444905805459438, 0.9064975009611688}
\definecolor{fau-med_dark-375}{rgb}{0.6235294117647059, 0.7667358708189158, 0.8597462514417532}
\definecolor{fau-med_dark-625}{rgb}{0.37254901960784315, 0.6112264513648596, 0.766243752402922}
\definecolor{fau-med_dark}{rgb}{0.0, 0.3803921568627451, 0.6274509803921569}
\definecolor{fau-med_light-125}{rgb}{0.9212610534409842, 0.9714571318723568, 0.9906497500961169}
\definecolor{fau-med_light-250}{rgb}{0.8425221068819685, 0.9429142637447135, 0.9812995001922338}
\definecolor{fau-med_light-375}{rgb}{0.7637831603229527, 0.9143713956170704, 0.9719492502883507}
\definecolor{fau-med_light-625}{rgb}{0.6063052672049212, 0.8572856593617839, 0.9532487504805844}
\definecolor{fau-med_light}{rgb}{0.37254901960784315, 0.7725490196078432, 0.9254901960784314}
\definecolor{fau-med}{rgb}{0.0, 0.6392156862745098, 0.8784313725490196}
\definecolor{fau-nat-125}{rgb}{0.9074817377931564, 0.9611226451364859, 0.8951787773933103}
\definecolor{fau-nat-250}{rgb}{0.8149634755863129, 0.922245290272972, 0.7903575547866205}
\definecolor{fau-nat-375}{rgb}{0.7224452133794694, 0.8833679354094579, 0.6855363321799308}
\definecolor{fau-nat-625}{rgb}{0.5374086889657823, 0.8056132256824298, 0.4758938869665513}
\definecolor{fau-nat-dark}{rgb}{0.13333333333333333, 0.5333333333333333, 0.2823529411764706}
\definecolor{fau-nat-light}{rgb}{0.5372549019607843, 0.803921568627451, 0.4745098039215686}
\definecolor{fau-nat_dark-125}{rgb}{0.8912418300653595, 0.941437908496732, 0.9099423298731257}
\definecolor{fau-nat_dark-250}{rgb}{0.782483660130719, 0.8828758169934641, 0.8198846597462515}
\definecolor{fau-nat_dark-375}{rgb}{0.6737254901960784, 0.8243137254901961, 0.7298269896193772}
\definecolor{fau-nat_dark-625}{rgb}{0.4562091503267973, 0.7071895424836601, 0.5497116493656287}
\definecolor{fau-nat_dark}{rgb}{0.13333333333333333, 0.5333333333333333, 0.2823529411764706}
\definecolor{fau-nat_light-125}{rgb}{0.9419300269127259, 0.9753940792003076, 0.9340561322568243}
\definecolor{fau-nat_light-250}{rgb}{0.8838600538254517, 0.9507881584006151, 0.8681122645136485}
\definecolor{fau-nat_light-375}{rgb}{0.8257900807381776, 0.9261822376009228, 0.8021683967704729}
\definecolor{fau-nat_light-625}{rgb}{0.7096501345636294, 0.8769703960015379, 0.6702806612841214}
\definecolor{fau-nat_light}{rgb}{0.5372549019607843, 0.803921568627451, 0.4745098039215686}
\definecolor{fau-nat}{rgb}{0.2627450980392157, 0.6901960784313725, 0.16470588235294117}
\definecolor{fau-phil-125}{rgb}{1.0, 0.9650595924644367, 0.8882891195693964}
\definecolor{fau-phil-250}{rgb}{1.0, 0.9301191849288735, 0.7765782391387928}
\definecolor{fau-phil-375}{rgb}{1.0, 0.8951787773933103, 0.6648673587081892}
\definecolor{fau-phil-625}{rgb}{1.0, 0.8252979623221838, 0.441445597846982}
\definecolor{fau-phil-dark}{rgb}{0.9098039215686274, 0.4666666666666667, 0.13333333333333333}
\definecolor{fau-phil-light}{rgb}{1.0, 0.8235294117647058, 0.44313725490196076}
\definecolor{fau-phil_dark-125}{rgb}{0.9886812764321414, 0.9330718954248366, 0.8912418300653595}
\definecolor{fau-phil_dark-250}{rgb}{0.977362552864283, 0.8661437908496732, 0.782483660130719}
\definecolor{fau-phil_dark-375}{rgb}{0.9660438292964244, 0.7992156862745098, 0.6737254901960784}
\definecolor{fau-phil_dark-625}{rgb}{0.9434063821607074, 0.6653594771241831, 0.4562091503267973}
\definecolor{fau-phil_dark}{rgb}{0.9098039215686274, 0.4666666666666667, 0.13333333333333333}
\definecolor{fau-phil_light-125}{rgb}{1.0, 0.9778546712802768, 0.9301191849288735}
\definecolor{fau-phil_light-250}{rgb}{1.0, 0.9557093425605536, 0.860238369857747}
\definecolor{fau-phil_light-375}{rgb}{1.0, 0.9335640138408304, 0.7903575547866205}
\definecolor{fau-phil_light-625}{rgb}{1.0, 0.889273356401384, 0.6505959246443676}
\definecolor{fau-phil_light}{rgb}{1.0, 0.8235294117647058, 0.44313725490196076}
\definecolor{fau-phil}{rgb}{1.0, 0.7215686274509804, 0.10980392156862745}
\definecolor{fau-tech-125}{rgb}{0.9330718954248366, 0.9527566320645905, 0.9635832372164552}
\definecolor{fau-tech-250}{rgb}{0.8661437908496732, 0.9055132641291811, 0.9271664744329104}
\definecolor{fau-tech-375}{rgb}{0.7992156862745098, 0.8582698961937716, 0.8907497116493657}
\definecolor{fau-tech-625}{rgb}{0.6653594771241831, 0.7637831603229527, 0.8179161860822761}
\definecolor{fau-tech-dark}{rgb}{0.2549019607843137, 0.4549019607843137, 0.5529411764705883}
\definecolor{fau-tech-light}{rgb}{0.6666666666666666, 0.7647058823529411, 0.8196078431372549}
\definecolor{fau-tech_dark-125}{rgb}{0.9064975009611688, 0.9315955401768551, 0.9438985005767013}
\definecolor{fau-tech_dark-250}{rgb}{0.8129950019223375, 0.8631910803537101, 0.8877970011534025}
\definecolor{fau-tech_dark-375}{rgb}{0.7194925028835064, 0.7947866205305651, 0.8316955017301038}
\definecolor{fau-tech_dark-625}{rgb}{0.532487504805844, 0.6579777008842753, 0.7194925028835064}
\definecolor{fau-tech_dark}{rgb}{0.2549019607843137, 0.4549019607843137, 0.5529411764705883}
\definecolor{fau-tech_light-125}{rgb}{0.9581699346405229, 0.970472895040369, 0.977362552864283}
\definecolor{fau-tech_light-250}{rgb}{0.9163398692810457, 0.9409457900807382, 0.954725105728566}
\definecolor{fau-tech_light-375}{rgb}{0.8745098039215686, 0.9114186851211072, 0.9320876585928489}
\definecolor{fau-tech_light-625}{rgb}{0.7908496732026143, 0.8523644752018454, 0.8868127643214149}
\definecolor{fau-tech_light}{rgb}{0.6666666666666666, 0.7647058823529411, 0.8196078431372549}
\definecolor{fau-tech}{rgb}{0.4666666666666667, 0.6235294117647059, 0.7098039215686275}
\definecolor{fau-wiso-125}{rgb}{0.9729334871203383, 0.8823836985774702, 0.8971472510572857}
\definecolor{fau-wiso-250}{rgb}{0.9458669742406767, 0.7647673971549405, 0.7942945021145713}
\definecolor{fau-wiso-375}{rgb}{0.918800461361015, 0.6471510957324107, 0.691441753171857}
\definecolor{fau-wiso-625}{rgb}{0.8646674356016917, 0.41191849288735105, 0.4857362552864283}
\definecolor{fau-wiso-dark}{rgb}{0.592156862745098, 0.10588235294117647, 0.1843137254901961}
\definecolor{fau-wiso-light}{rgb}{0.8627450980392157, 0.4117647058823529, 0.48627450980392156}
\definecolor{fau-wiso_dark-125}{rgb}{0.9488196847366398, 0.8877970011534025, 0.8976393694732795}
\definecolor{fau-wiso_dark-250}{rgb}{0.8976393694732795, 0.7755940023068051, 0.795278738946559}
\definecolor{fau-wiso_dark-375}{rgb}{0.8464590542099193, 0.6633910034602075, 0.6929181084198386}
\definecolor{fau-wiso_dark-625}{rgb}{0.7440984236831988, 0.4389850057670127, 0.48819684736639757}
\definecolor{fau-wiso_dark}{rgb}{0.592156862745098, 0.10588235294117647, 0.1843137254901961}
\definecolor{fau-wiso_light-125}{rgb}{0.9827758554402153, 0.9261822376009228, 0.9355324875048059}
\definecolor{fau-wiso_light-250}{rgb}{0.9655517108804306, 0.8523644752018454, 0.8710649750096117}
\definecolor{fau-wiso_light-375}{rgb}{0.948327566320646, 0.7785467128027681, 0.8065974625144176}
\definecolor{fau-wiso_light-625}{rgb}{0.9138792772010765, 0.6309111880046137, 0.6776624375240292}
\definecolor{fau-wiso_light}{rgb}{0.8627450980392157, 0.4117647058823529, 0.48627450980392156}
\definecolor{fau-wiso}{rgb}{0.7843137254901961, 0.06274509803921569, 0.1803921568627451}
\usetikzlibrary{arrows.meta}
\tikzset{%
  >={Latex[width=2mm,length=2mm]},
            base/.style = {rectangle, rounded corners, draw=black,
                           minimum width=4cm, minimum height=1cm,
                           text centered, font=\sffamily},
  activityStarts/.style = {base, fill=fau-fau_light-375, minimum width =\textwidth*0.1, minimum height = 0.5cm},
    activityStarts2/.style = {base, fill=fau-fau_light-125, minimum width =\textwidth*0.1, minimum height = 0.5cm},
       startstop/.style = {base, fill=fau-fau_light-625, minimum width =\textwidth*0.1},
         process/.style = {base, minimum width = 0.2cm, minimum height = 0.2cm, draw = white},
    intermediate/.style = {base, minimum width = 0.2cm, minimum height = 0.2cm},
}
\tikzset{
    *|/.style={
        to path={
            (perpendicular cs: horizontal line through={(\tikztostart)},
                                 vertical line through={(\tikztotarget)})
            -- (\tikztotarget) \tikztonodes
        }
    }
}
\tikzset{
    |*/.style={
        to path={
            (perpendicular cs: vertical line through={(\tikztotarget)})
            -- (\tikztotarget) \tikztonodes
        }
    }
}

\begin{document}
\bstctlcite{IEEEexample:BSTcontrol}



\title{Contributing Components of Metabolic Energy Models to Metabolic Cost Estimations in Gait}
\author{Markus Gambietz, Marlies Nitschke, J\"org Miehling, Anne D. Koelewijn

\thanks{This work was funded by the Deutsche Forschungsgemeinschaft (DFG, German Research Foundation) – SFB 1483 – Project-ID 442419336, EmpkinS.}

\thanks{Markus Gambietz, Marlies Nitschke, and Anne D. Koelewijn are with the Machine Learning and Data Analytics Lab, Department Artificial Intelligence in Biomedical Engineering (AIBE), Faculty of Engineering, Friedrich-Alexander-Universit\"{a}t Erlangen-N\"{u}rnberg, 91052 Erlangen, Germany (email: markus.gambietz@fau.de)}
\thanks{J\"org Miehling is with the Chair of Engineering Design, Department of Mechanical Engineering, Faculty of Engineering, Friedrich-Alexander-Universit\"{a}t Erlangen-N\"{u}rnberg, 91058 Erlangen, Germany}}


\maketitle

\begin{abstract}
\textit{Objective:} As metabolic cost is a primary factor influencing humans' gait, we want to deepen our understanding of metabolic energy expenditure models. Therefore, this paper identifies the parameters and input variables, such as muscle or joint states, that contribute to accurate metabolic cost estimations. 
\textit{Methods:} We explored the parameters of four metabolic energy expenditure models in a Monte Carlo sensitivity analysis. Then, we analysed the model parameters by their calculated sensitivity indices, physiological context, and the resulting metabolic rates during the gait cycle. The parameter combination with the highest accuracy in the Monte Carlo simulations represented a quasi-optimized model. In the second step, we investigated the importance of input parameters and variables by analysing the accuracy of neural networks trained with different input features. 
\textit{Results:} Power-related parameters were most influential in the sensitivity analysis and the neural network-based feature selection.
We observed that the quasi-optimized models produced negative metabolic rates, contradicting muscle physiology. Neural network-based models showed promising abilities but have been unable to match the accuracy of traditional metabolic energy expenditure models.
\textit{Conclusion:} We showed that power-related metabolic energy expenditure model parameters and inputs are most influential during gait. Furthermore, our results suggest that neural network-based metabolic energy expenditure models are viable. However, bigger datasets are required to achieve better accuracy.
\textit{Significance:} As there is a need for more accurate metabolic energy expenditure models, we explored which musculoskeletal parameters are essential when developing a model to estimate metabolic energy.

\end{abstract}


\section*{Nomenclature}
\addcontentsline{toc}{section}{Nomenclature}
\begin{IEEEdescription}[]
\item[ANN]  \hspace{0.5cm} Artificial Neural Network
\item[CMC]  \hspace{0.5cm} Coefficient of Multiple Correlation
\item[KIMR15]  \hspace{0.5cm} MEE Model by Kim et al.
\item[KSstat]   \hspace{0.5cm} Kolmogorov–Smirnov Statistic
\item[LICH05]  \hspace{0.5cm} MEE Model by Lichtwark et al.
\item[LOO]  \hspace{0.5cm} Leave-One-Out
\item[MARG68]  \hspace{0.5cm} MEE Model by Margaria
\item[MC]   \hspace{0.5cm} Monte Carlo
\item[MEE]  \hspace{0.5cm} Metabolic Energy Expenditure
\item[MINE97]  \hspace{0.5cm} MEE Model by Minetti et al.
\item[RMC]  \hspace{0.5cm} Repeated Measures Correlation
\item[RMSE] \hspace{0.5cm} Root Mean Square Error
\end{IEEEdescription}
%
\IEEEpeerreviewmaketitle

\section{Introduction}

Over the last decades, improvements in hard- and software, as well as an improved understanding of gait, have enabled researchers to design smart active wearable devices, such as prostheses and exoskeletons. Those devices can improve gait and, thereby, participation in society for persons with a gait disability. One of the ways these devices improve participation in society is by reducing the energy consumption or metabolic cost of walking.
Since the middle of the last century, experimental research has shown that people choose many gait variables, such as speed, cadence and stride length, to reduce the metabolic cost of gait~\cite{Ralston.17, Zarrugh.1974}. Furthermore, it was recently shown that even in minutes to hours, persons would adjust their gait to perform this optimization~\cite{selinger.2015, https://doi.org/10.1113/jphysiol.2012.245506}. 
Therefore, there is a need for accurate assessment of metabolic energy consumption.

Metabolic energy expenditure (\acrshort{mee}) can be assessed computationally via MEE models and experimentally, the latter of which is more commonly used. These experiments use direct or indirect calorimetry. In direct calorimetry, the heat produced by the human body is measured directly, for example, in a heat chamber~\cite{levine_2005}. The system setup and operation are complex and costly~\cite{levine_2005}. Therefore, indirect calorimetry is used more frequently. In indirect calorimetry, the energy expenditure is determined by measuring the amount of inhaled and exhaled oxygen and carbon dioxide~\cite{Ferrannini.1988}. Oxygen consumption is directly related to energy expenditure when the performed exercise is aerobic. However, also indirect calorimetry has limitations. First, it requires about 3 minutes to adjust to the level of the oxygen inhalation required for the exercise~\cite{doi:10.1152/japplphysiol.00445.2014}. Secondly, breathing is variable; consequently, to get an accurate estimate of energy expenditure based on the average over several breaths, a more extended measurement period is required. Therefore, participants must perform the exercise of interest for at least 5 minutes, though longer periods could improve accuracy. Hence, indirect calorimetry estimates the averaged MEE over the measurement period and does not provide instantaneous MEE estimates at a specific point in time. Furthermore, direct and indirect calorimetry can only provide full-body energy expenditure, including exercise and the expenditure of the brain, the gastrointestinal system, and other organs. Therefore, it is not straightforward to calculate the "net metabolic cost" or the metabolic cost of the activity itself, and the contribution of different muscles cannot be individually assessed.

Instead, MEE models could be used to estimate instantaneous energy expenditure at the muscle level~\cite{Miller.2014, Koelewijn.2019}. MEE models calculate energy expenditure based on kinetic and kinematic movement variables. Margaria~\cite{Margaria.1968} was the first to describe a relationship between eccentric and concentric muscle contraction based on muscle efficiency during shortening and lengthening. Further MEE models tried to form new approaches for the computational determination of MEE. They were based on, for example, human gait in two dimensions (2D)~\cite{Minetti.1997}, three dimensions (3D)~\cite{Kim.2015}, or the contraction of the sartorius muscle of a frog~\cite{Bhargava.2004}. Depending on the approach, MEE models vary in complexity and contain between two and fifteen empirical parameters, while the amount of parameters required to represent muscle physiology accurately remains unclear. 

So far, comparative studies have focused on the accuracy and correlation of the \acrshort{mee} models compared to measured metabolic rates. Miller et al.~\cite{Miller.2014} compared five MEE models at self-chosen walking speed and cadence. In the study by Koelewijn et al.~\cite{Koelewijn.2019}, the models from~\cite{Miller.2014}, and two additional MEE models were tested for their responses in altered environments, e.g., different walking speeds and inclines. Both studies reported an underestimation of metabolic cost in gait~\cite{Miller.2014, Koelewijn.2019}. A recent study aimed to estimate changes in energy expenditure in older people due to changes in walking patterns using MEE models. However, the MEE models could only explain about one-third of the measured changes~\cite{Pimetel4}. Therefore, the accuracy of MEE models should be improved to predict changes in the metabolic rate correctly.

Furthermore, the use of MEE models in simulations of human gait has proven to be successful in replicating natural walking patterns by minimizing energy-related objectives~\cite{BERTRAM2001445} with the use of both simple~\cite{Srinivasan200672} and complex musculoskeletal models~\cite{Miller.2014}. Muscle-space MEE models have been used as the objective~\cite{koelewijn_metabolic_2018,falisse_rapid_2019} for muscle-driven simulations. For torque-driven simulations, a joint-space MEE model \cite{Kim.2015} could be used as the objective. In such gait simulations, the objective function's complexity significantly impacts the computational time~\cite{koelewijn_metabolic_2018}. Hence, from a simulation standpoint, there is a motivation to use accurate MEE models with lower complexity to enhance computational efficiency.

In addition to the challenges associated with the accuracy of MEE models, there are other important considerations to address. In biomechanical processing, musculoskeletal models often rely on cadaveric measurements of elderly individuals as a basis \cite{Zatsiorsky}. However, when scaling these generic musculoskeletal models, crucial muscle parameters required for accurate calculation of metabolic costs are not personalized \cite{Delp.2007}. Moreover, muscle force estimation is resource intensive and depends on the specific optimization problem formulation~\cite{Groote.2016}. Consequently, the calculation of metabolic costs is already based on inherently inaccurate data. An alternative approach to mitigate some of the limitations at the muscle level is to estimate metabolic costs at the joint level~\cite{Kim.2015}. Furthermore, metabolic cost models cannot be fully validated since energy expenditure cannot be measured at a muscle level during gait.

\begin{figure}
    \centering
    \includegraphics[width=\columnwidth]{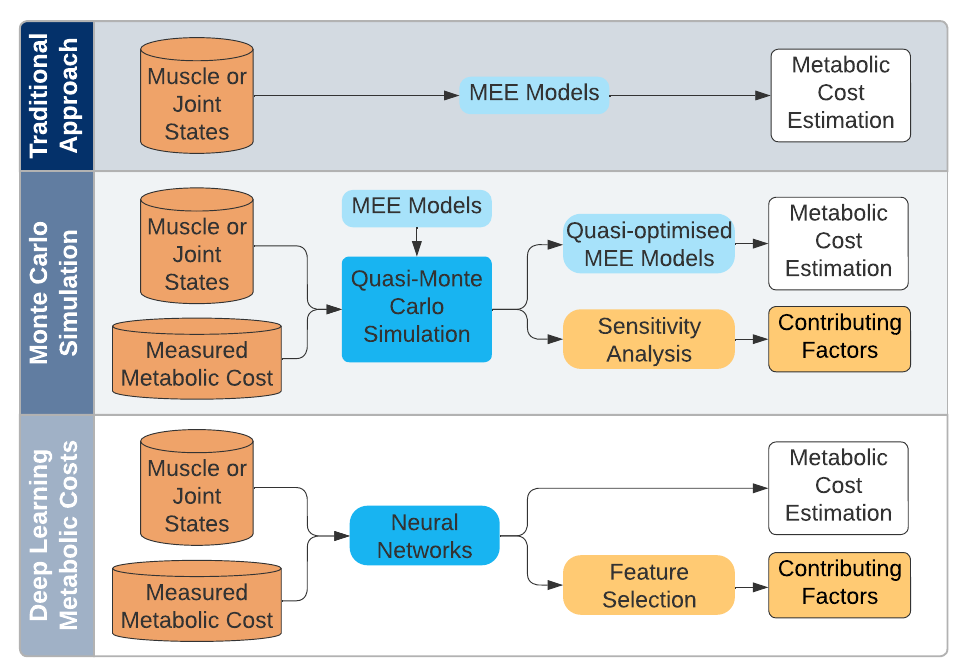}
    \caption{Overview of the approach in this paper. {We identified the contributing factors in current MEE models locally via a sensitivity analysis. Furthermore, we searched for contributing components globally by performing a feature selection using neural networks.}}
    \label{fig:Flowchart}
\end{figure}

In this study, we aim to deepen our understanding of MEE estimations by investigating the importance of different model parameters both locally and globally (see Figure \ref{fig:Flowchart}). We investigate the sensitivity of model parameters of the different MEE models to understand which parameters are essential when developing more accurate MEE models. First, we undertook a local sensitivity analysis by performing a Monte Carlo simulation on the empirical model parameters to determine their influence quantitatively. Furthermore, we identified the MEE models with the highest accuracy in the Monte Carlo simulation and investigated how the MEE estimations of these models differ from the original version.
In the second step, we searched for the globally most optimal feature set. {Therefore, we used neural networks as purely statistical representations for metabolic models and trained them with different input modalities.}


\section{Methods}

\subsection{Data set}\label{chap:dataset}
We have used all available trials of the publicly available data set of Koelewijn et al.~\cite{Koelewijn.2018}.
It comprises six female and six male participants, averaging \SI{24(5)} years in age, \SI{173(8)}{\cm} in height 
and a weight of \SI{70(12)}{\kg}. Gait of each participant was recorded at two velocities (\SI{0.8}{\meter\per\second}, \SI{1.3}{\meter\per\second}) and three inclines (\SI{-8}{\percent}, \SI{0}{\percent}, \SI{+8}{\percent}). 


The data set contains processed data consisting of joint angles, joint velocities, and joint moments, of the hip, knee and ankle in the sagittal plane, as well as muscle activations, stimulations, and lengths, for eight muscles in each leg. The data processing was described and performed by Koelewijn et al. \cite{Koelewijn.2019}. Koelewijn et al. estimated joint angles, joint velocities, and joint moments using a 2D link-segment leg model. Then, they used a dynamic optimization procedure \cite{Groote.2016} to estimate muscle activations, stimulations, and length. Furthermore, the average metabolic cost for each trial, measured through indirect calorimetry, was reported~\cite{Koelewijn.2018}. This data can be used directly as inputs for the \acrshort{mee} models described in section \ref{chap:meemodels}. 

\subsection{Metabolic energy expenditure models}\label{chap:meemodels}
In our sensitivity analysis, we compared and analyzed all models from Koelewijn et al.'s study \cite{Koelewijn.2019}. We present the MEE models MARG68~\cite{Margaria.1968}, MINE97~\cite{Minetti.1997}, LICH05~\cite{Lichtwark.2005}, and KIMR15~\cite{Kim.2015} in the manuscript, while the MEE models by Bhargava et al.~\cite{Bhargava.2004}, Houdijk et al.~\cite{Houdijk.2006}, and Umberger et al.~\cite{Umberger.2003} are added in the supporting document, since these models are conceptionally similar to LICH05.
MARG68, MINE97, and LICH05 use muscle states, while KIMR15 uses joint states to estimate the instantaneous metabolic rate $\dot{E}$ [\si{\watt}]. The metabolic cost $C_{calc}$ can be derived from the metabolic rate~\cite{Koelewijn.2019}:
\begin{equation}
    C_{calc} = \frac{1}{Tmv} \int_{t=0}^T \sum_{i=0}^{n_{mus}} \dot{E}_i\text{d}t,
\end{equation}
where $T$ is the duration of the motion, $m$ the body mass, $v$ the velocity, and $n_{mus}$ the number of muscles $i$ for which the metabolic rate $\dot{E}$ is summed.

\subsubsection*{MARG68}
The model by Margaria \cite{Margaria.1968} is based on the observation of differences in MEE between muscle lengthening and shortening. Margaria estimated that the shortening efficiency $\eta_s$ is \SI{25}{\percent}, with the lengthening efficiency $\eta_l$ being \SI{120}{\percent} in relation to the power $\dot{w}$, depending on the sign of the fibre velocity $v_{ce}$:
\begin{equation}\label{for:marg}
    \dot{E} = 
    \begin{cases}
    \frac{\dot{w}}{{\eta_s}} & v_{ce} < 0    \\
    - \frac{\dot{w}}{{\eta_l}} & v_{ce} \geq 0.
    \end{cases}
\end{equation}
The power $\dot{w}$ is the product of muscle velocity $v_{ce}$ and muscle contractile element force $F_{ce}$, which itself is the product of maximum isometric force $F_{max}$, activation $a$, and the Hill-type force-length $f_{FL}(\Tilde{l}_{ce})$ and force-velocity $f_{FV}(\Tilde{v}_{ce})$ relationships.

\subsubsection*{MINE97}
The model of Minetti and Anderson \cite{Minetti.1997} estimates the metabolic rate $\dot{E}$ from the shortening velocity based on an empirical relationship:
 \begin{equation}\label{for:MINE97}
     \dot{E} = a v_{max} F_{max} \phi(v_{ce}),
\end{equation}
where $a$ is the muscle activation, $v_{max}$ the maximum muscle fibre velocity, and $F_{max}$ the maximum isometric force. $\phi$ is a relationship between fibre velocity and energy expenditure, with $\Bar{v}_{ce}=\Tilde{v}_{ce}/\Tilde{v}_{max}$ being the relation of the normed muscle fibre velocity $\Tilde{v}_{ce}$ to the maximum muscle fibre velocity $v_{max}$.:
\begin{equation}\label{for:MINE972}
    \phi(v_{ce}) = \frac{p_{mi,1}+p_{mi,2}\Bar{v}_{ce}+p_{mi,3}\Bar{v}_{ce}^2}{p_{mi,4}+p_{mi,5}\Bar{v}_{ce}+p_{mi,6}\Bar{v}_{ce}^2+p_{mi,7}\Bar{v}_{ce}^3}.
\end{equation}
The parameters values $p_{mi,1...7}$ of the original model are [$0.054$, $0.506$, $2.46$, $1$, $-1.13$, $12.8$, $-1.64$].
\subsubsection*{LICH05}
In the model by Lichtwark et al. \cite{Lichtwark.2005}, the total energy rate $\dot{E}$ is calculated through the sum of power $\dot{w}$ and heat rate $\dot{h}$ produced due to muscle shortening and lengthening as well as maintenance of the activation. In this context, the parameters $p_{l,1...8}$ represent the unnamed empirical parameters that are present in the model.
The maintenance heat rate is dependent on a decay function $\gamma$ and the fibre velocity where shortening is defined as positive:
\begin{multline}
    \dot{h}_M = \gamma \frac{\Tilde{v}_{max}}{G^2} \cdot \\
    \begin{cases}
        1 & v_{ce} > 0 \\
        \left(p_{l,2}+(1-p_{l,2})e^{-p_{l,3}\Tilde{v}_{ce}(f_{FV}(\Tilde{v}_{ce})-1)}\right) &v_{ce}  \leq 0,
    \end{cases}
\end{multline}
where $f_{FV}(\Tilde{v}_{ce})$ is the Hill-type force-velocity characteristic, $G$ is the curvature of $f_{FV}(\Tilde{v}_{ce})$, and $\gamma$ is defined as:
\begin{equation}
    \gamma= p_{l,4} e^{- p_{l,5}t_{stim}} + p_{l,6} e^{- p_{l,7}t_{stim}},
\end{equation}
where $t_{stim}$ is the duration for which the activation of a muscle has been higher than a threshold $t_{act}$.
The shortening-lengthening heat rate is defined as:
\begin{equation}
    \dot{h}_{SL} = 
    \begin{cases}
        \frac{\Tilde{v}_{ce}}{G} & v_{ce} > 0 \\
        -p_{l,8} f_{FV}(\Tilde{v}_{ce})\Tilde{v}_{ce}&v_{ce}  \leq 0.
    \end{cases}
\end{equation}

A modified version \cite{Lichtwark.2007} of the original LICH05 model \cite{Lichtwark.2005} was used, where the overall heat rate is defined as:
\begin{multline}
    \dot{h} = aF_{max} \left( p_{l,1} \dot{h}_M + f_{FL}(\Tilde{l}_{ce})\left((1-p_{l,1}) \dot{h}_M  +\dot{h}_{SL}\right) \right).
\end{multline}
The empirical parameters of LICH05 were the following: $G=4$, $t_{act}=0.1$, and $p_{l,1...8}=$ [$0.3$, $0.3$, $7$, $0.8$, $0.72$, $0.175$, $0.022$, $0.5$].

\subsubsection*{KIMR15}
The model created by Kim et al. \cite{Kim.2015} is the only joint-space \acrshort{mee} model and calculates the metabolic rate based on joint angles $q$, joint velocities $\dot{q}$, and joint moments $M$:
\begin{equation}
    \dot{E} = \dot{h}_M|\dot{q}|_{max} |M| + \dot{h}_{SL}|M\dot{q}| + \dot{q}_{cc}(M\dot{q})_{max} + M\dot{q},
\end{equation}
where $\dot{h}_M = 0.054$, $\dot{h}_{SL,s} = 0.283$ for shortening and $\dot{h}_{SL,l} = 1.423$ for lengthening, respectively, and $\dot{q}_{cc} = 0.004$ is the co-contraction heat rate.
%
%
%
\subsection{Quasi-Monte Carlo sensitivity analysis}\label{chap:MCmethod}


We conducted a sensitivity analysis to determine which parameters in the MEE models have a large or small influence on the calculated metabolic cost in gait. 
For this purpose, we used a quasi-Monte Carlo (MC) simulation by drawing a large number of random MEE model parameter sets, assessing the error and correlation to the measured metabolic cost for every randomised parameter set. The random parameter sets include all empirical parameters in the respective MEE models. An overview of the respective search spaces per parameter is listed in the supporting document.

In each iteration, each gait trial's metabolic cost was calculated using a quasi-randomly parametrised MEE model. The MEE model parameters were sampled using Saltelli's extension of the Sobol sequence, as implemented in PyTorch \cite{Sobol.2001, Saltelli.2010, Paszke.03.12.2019}. The quasi-random Sobol sequences enable the generation of low-discrepancy sequences, which cover the space of possible parameter combinations more evenly than pseudo-random methods. As a metric, the root-mean-squared error (RMSE) and the repeated measures correlation (RMC)~\cite{Bakdash.2017} between the calculated and measured metabolic cost were determined. The RMSE is an indicator for the absolute accuracy of the MEE models, while the RMC is an indicator for intra-subject correlation. Furthermore, the coefficient of multiple correlation (CMC)~\cite{Ferrari.2010} was used to assess the shape similarity of metabolic rate curves between the MEE models.

After sampling the model parameters and calculating the corresponding RMSE, the sensitivity indices $S_{i}$ were computed via the Kolgomorov-Smirnoff statistic (KSstat) \cite{Pianosi.2016} with the scipy library's "kstest" \cite{2020SciPy-NMeth}.
Extensive testing showed that a combination of $10^5$ model evaluations with a behavioural set size of $10^2$ was suitable for both exploring the parameter space and calculating the KSstat. These sensitivity indices can be used to quantitatively rank the individual parameters according to their contribution to the metabolic cost estimation. We enhanced the understanding of simplest models, MARG68 and MINE97, by incorporating visual representations to illustrate the Monte-Carlo simulation. This was not possible for the other models due to the high number of parameters.

Furthermore, the model parameter sets with the lowest RMSE values in the MC simulation resulted in quasi-optimised MEE models since those estimated the measured metabolic cost best. We performed a subject-wise leave-one-out (LOO) cross-validation to report the RMSE and RMC achieved with the quasi-optimised MEE models.






\subsection{Deep MEE Models}\label{chap:DLmethod}
We trained artificial neural networks (ANNs) to estimate MEE using different feature combinations of muscle and joint parameters and variables as input to make statements about the necessity of these features. Furthermore, we determined whether ANNs can outperform the classical MEE models on our data set. Like the MEE models shown in section II-B, we trained the ANNs with combinations of muscle or joint states as an input vector to return metabolic rate per muscle or joint at each time point. 

\begin{figure}[]
    \centering
    \includegraphics[scale=0.5]{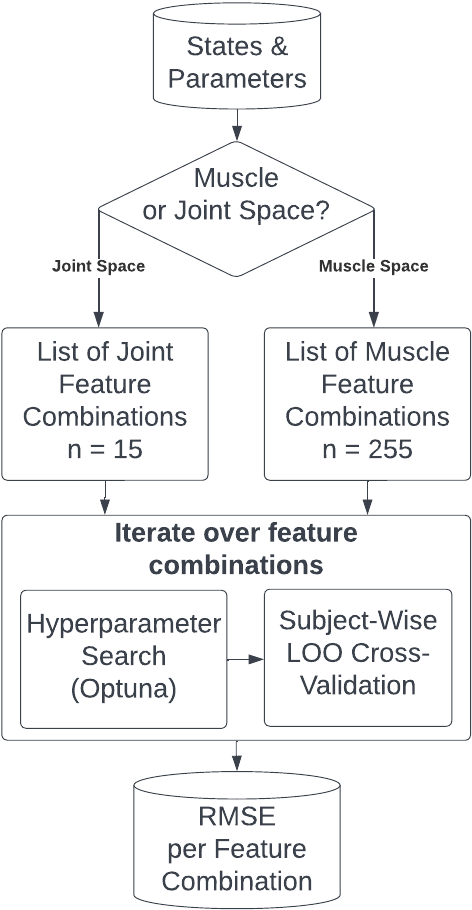}
    \caption{Deep-Learning MEE model pipeline.}
    \label{fig:InputVectorsOver}
\end{figure}

{ANNs are a purely statistical model that does not take prior knowledge of the biological system into account, allowing us to search for feature combinations globally.} As shown in Figure \ref{fig:InputVectorsOver}, every combination of the input features used in the different MEE models was tested to assess which feature combinations, in both joint and muscle space, are crucial for \acrshort{mee} predictions. In the muscle space, the eight input features are normalised fibre length $\Tilde{l}_{ce}$, normalised fibre velocity $\Tilde{v}_{ce}$, muscle activation $a$, maximum isometric force $F_{max}$, fast-twitch fibre percentage $r_{ft}$, muscle width $w_h$, optimal fibre length $l_{ce,opt}$, and muscle stimulation $e$.
The joint-space ANNs received feature combinations of joint angles $q$, velocities $\dot{q}$, accelerations $\ddot{q}$, and moments $M$. Each input was scaled to a minimum / maximum value between $0$ and $1$.

The ANNs estimate the metabolic rate of each joint or muscle at each time point, as the traditional metabolic models do. After the forward pass, the metabolic rate is summed over all muscles or joints and then averaged over the gait cycle. Then, the loss is formed by comparing the estimated average metabolic rate to the gait cycle's average metabolic rate computed from the measured metabolic cost. Therefore, our training approach can be classified as inexact supervised learning since the metabolic rate for each muscle and time point is indirectly supervised \cite{Zhou10.1093/nsr/nwx106}. We used this approach to maintain the ability of MEE models to estimate individual muscle or joint metabolic rates and to be able to investigate the metabolic rate over time.


All models were set up as dense networks in PyTorch~\cite{Paszke.03.12.2019, BaiTCN2018}, where a combination of the mean-squared error loss function and Adam as the optimiser~\cite{Kingma.22.12.2014} proved most reliable. ReLU and Leaky ReLU activation functions were used~\cite{Nair.2010, Goodfellow.2016, Xu.05.05.2015}.
Hyperparameter tuning was done using Optuna's TPESampler~\cite{Akiba.25.07.2019}, with tunable hyperparameters being the number of layers, layer sizes, batch size, learning rate, and weight decay.

The feature combinations were evaluated by comparing their respective RMSE after subject-wise LOO cross validation. The training set of each fold, which contains the trials from the remaining eleven subjects, was randomly split into \SI{70}{\percent} training and \SI{30}{\percent} test set. For each feature combination, hyperparameter tuning was performed during the first cross-validation step. The networks were trained on a cluster equipped with NVIDIA RTX 2080s.
\section{Results}
\label{chap:results}

\begin{figure*}[t!]
    \begin{minipage}[t]{0.6\linewidth}
    \textbf{MARG68} \\
    \hfill
    \vspace{0.05cm}

    \includegraphics[]{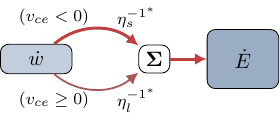} \\ \\
    \textbf{KIMR15} \\ 
        \vspace{0.05cm}

         \includegraphics[]{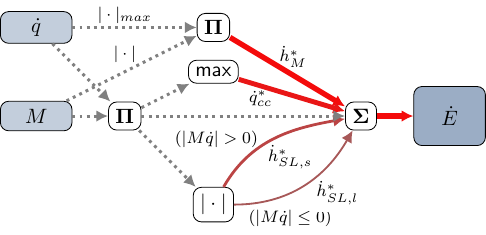}
    \end{minipage}\hfill
    \begin{minipage}[t]{0.39\linewidth}
        \textbf{MINE97} \\
            \vspace{0.1cm}

                 \includegraphics[]{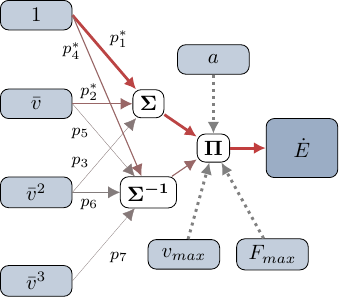}

    \end{minipage}\hfill
    \vspace{0.2cm}

    \begin{minipage}[t]{1\linewidth}
        \vspace{0.5cm}
        
        \textbf{LICH05}
        
    \end{minipage}
         \begin{minipage}[t]{0.98\linewidth}
         \centering
         \includegraphics[]{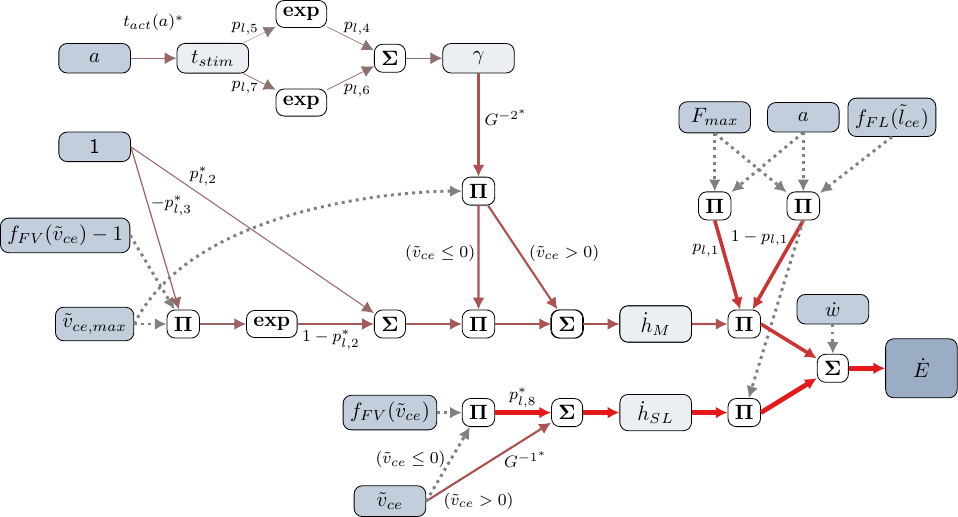}
     \end{minipage}
    \caption{These flowcharts illustrate the computation steps of the respective MEE models and the sensitivity associated with the parameters at each computation step. The thicker red arrows indicate higher sensitivity indices, while the thinner gray arrows indicate lower sensitivity indices. Parameters with significant sensitivity indices ($p<0.05$) are marked with an asterisk. The gray dotted arrows represent input propagation without modifications. When multiple arrows end at the same node, the brighter one is used for continuation. The metabolic rate $\Dot{E}$ is presented in dark blue boxes, while inputs are shown in blue, and intermediate results (refer to Chapter \ref{chap:meemodels}) are shown in light blue boxes.}
         \label{fig:glowchart}
\end{figure*}
    
\subsection{Sensitivity analysis on MEE model parameters}\label{chap:sensochad}
The identification of the most sensitive parameters is model-dependent (see Figure \ref{fig:glowchart}). In the case of MINE97 and KIMR15, the parameters associated with force ($p_{mi,1}$ in MINE97 and $\dot{h}_m$ in KIMR15) exhibit the highest sensitivity. Conversely, in MARG68 and LICH05, the parameters associated with shortening exhibit the highest sensitivity indices. In KIMR15, the shortening parameter also outranks the lengthening parameter in sensitivity.
A complete list of sensitivity indices can be found in the supporting document. However, note that the indices cannot be compared between the MEE models because each is based on a separate sensitivity analysis.

\begin{figure*}[h!]
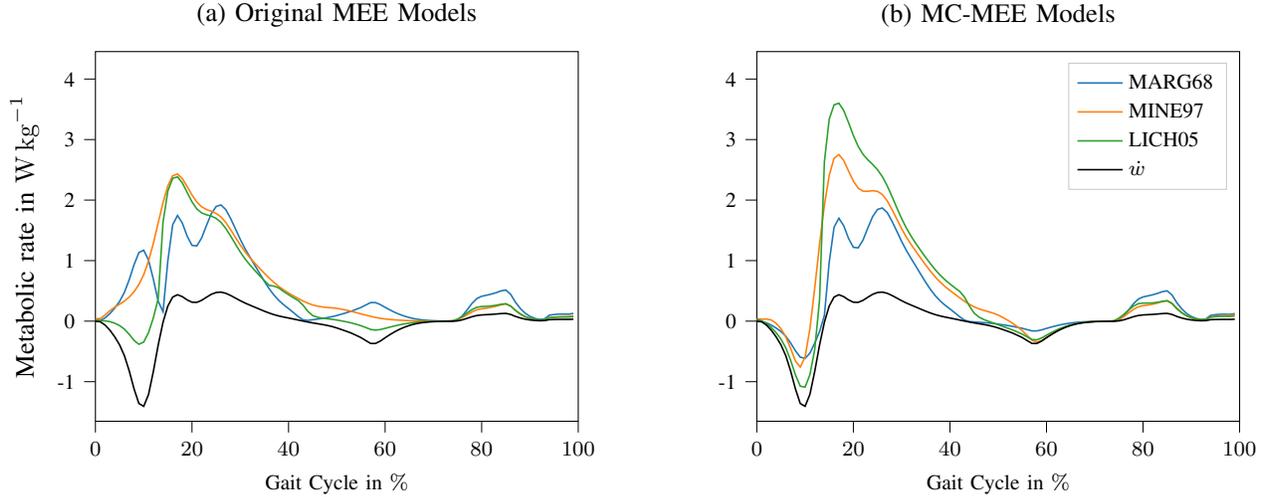

     \begin{minipage}[t]{0.48\linewidth}
         \centering
         \begin{tikzpicture}
\definecolor{color0}{rgb}{0.12156862745098,0.466666666666667,0.705882352941177}
\definecolor{color1}{rgb}{1,0.498039215686275,0.0549019807843137}
\definecolor{color2}{rgb}{0.172549019807843,0.627450980392157,0.172549019807843}
\definecolor{color3}{rgb}{0.83921568627451,0.152941176470588,0.156862745098039}
\begin{groupplot}[group style={group size=1 by 1,  horizontal sep=1.2cm, vertical sep=1.3 cm}]
\nextgroupplot[
title = {(a) Original MEE Models},
tick align=outside, height = 6.5cm, width = 8cm,
tick pos=left,
x grid style={white!69.0198078431373!black},
xtick style={color=black, font=\footnotesize},
y grid style={white!69.0198078431373!black},
ymin=-130,ymax=350,xmin=-0,xmax=100,
ytick={-100,0,100,200,300},
ylabel style={align=center},
legend cell align={left},
ylabel={Metabolic rate in \si{\watt\per\kilo\gram}},
legend style={fill opacity=0.8, draw opacity=1, text opacity=1, at={(0.97,0.03)}, anchor=south east, draw=white!80!black, font = \footnotesize},
ytick style={color=black, font=\footnotesize},
xlabel={\footnotesize Gait Cycle in \si{\percent}},
tick label style = {font=\footnotesize},
ytick={-78.6,0,78.6,157.2,235.8,314.4},
yticklabels={-1,0,1,2,3,4}
]
\input{plots/all/VastiFalse}

\end{groupplot}

\end{tikzpicture}
     \end{minipage}
     \hspace{0.5pt}
     \centering
     \begin{minipage}[t]{0.48\linewidth}
         \centering
         \begin{tikzpicture}
\definecolor{color0}{rgb}{0.12156862745098,0.466666666666667,0.705882352941177}
\definecolor{color1}{rgb}{1,0.498039215686275,0.0549019807843137}
\definecolor{color2}{rgb}{0.172549019807843,0.627450980392157,0.172549019807843}
\definecolor{color3}{rgb}{0.83921568627451,0.152941176470588,0.156862745098039}
\begin{groupplot}[group style={group size=2 by 4,  horizontal sep=1.2cm, vertical sep=1.3 cm}]
\nextgroupplot[
title = {(b) MC-MEE Models},
tick align=outside, height = 6.5cm, width = 8cm,
tick pos=left,
x grid style={white!69.0198078431373!black},
xtick style={color=black, font=\footnotesize},
y grid style={white!69.0198078431373!black},
ymin=-130,ymax=350,xmin=-0,xmax=100,
ylabel style={align=center},
legend cell align={left},
ytick={-100,0,100,200,300},
legend style={fill opacity=0.8, draw opacity=1, text opacity=1, at={(0.974,0.97)}, anchor=north east, draw=white!80!black, font = \footnotesize},
ytick style={color=black, font=\footnotesize},
xlabel={\footnotesize Gait Cycle in \si{\percent}},
tick label style = {font=\footnotesize},
ytick={-78.6,0,78.6,157.2,235.8,314.4},
yticklabels={-1,0,1,2,3,4}
]
\input{plots/all/VastiTrue}
\addlegendentry{MARG68}
\addlegendentry{MINE97}
\addlegendentry{LICH05}

\addlegendentry{$\Dot{w}$};

\end{groupplot}

\end{tikzpicture}
     \end{minipage}
     \caption[MEE curves from the muscle-space models]
     {A representative participant's vasti muscular metabolic rate according to the muscle-space MEE models in level walking. Left: Original MEE models. Right: Best-performing quasi-optimised MEE models from the MC simulation. Power $\dot{w}$ of the muscle annotated in black.}
     \label{fig:allplots}
\end{figure*}

After performing the quasi-Monte Carlo simulation, the quasi-optimised MEE models better fit the measured metabolic cost than the original MEE models since the RMSE improved on average by \SI{44(13)}{\percent} and the RMC increased slightly (see Table \ref{tab:overallp}). 
Figure \ref{fig:allplots} shows examplary metabolic rate curves the all muscle-space MEE models.
Between the curves of the original MEE models, a CMC of $0.84$ was observed. The quasi-optimised MEE models show a CMC of $0.92$ between each other, meaning the similarity between the curves is more prominent in the quasi-optimised MEE models, and they are all optimised towards the same shape. In the following, the individual models are described and pictured. 

\begin{table}[h!]
    \centering
    \caption[Overview of all \acrshort{mee} model performances.]{
    Overview of all repeated measures correlations (RMC) and root mean square errors (RMSE) for the original and quasi-optimised MEE models. The RMC and RMSE were obtained from the Monte Carlo simulation using a leave-one-out cross-validation.} \begin{tabular}{c|c|c|c|c}
     & \multicolumn{2}{c|}{Original} & \multicolumn{2}{c}{Quasi-Optimised MEE}\\
    Model & RMC & \acrshort{rmse} & RMC & \acrshort{rmse} \\
    &&[\si{\joule\per\kilo\gram\per\meter}]&&[\si{\joule\per\kilo\gram\per\meter}]\\
    \hline 
    MARG68 & $0.90$ & $1.19$ & $0.92$ & $0.79$ \\
    MINE97 & $0.95$ & $1.00$ & $0.93$ & $0.73$ \\
    LICH05 & $0.96$ & $1.06$ & $0.96$ & $0.68$ \\
    KIMR15 & $0.91$ & $1.98$ & $0.89$ & $1.15$ \\
    \end{tabular}
    \label{tab:overallp}
\end{table}


\subsubsection*{MARG68}
\begin{figure}[h!]
    \centering
    \begin{tikzpicture}
\begin{groupplot}[group style={group size=2 by 2,  horizontal sep=1.50cm, vertical sep=5 cm}, scale = 0.4, /tikz/background rectangle/.style={draw=thick}]
\nextgroupplot[
tick align=outside, height = 6.5cm, width = 6.5cm, 
tick pos=left,
x grid style={white!69.0196078431373!black},
xtick style={color=black},
y grid style={white!69.0196078431373!black},
xlabel={\footnotesize $\eta_s$},
ylabel={\footnotesize $\eta_l$},
ymin=-5,ymax=5,xmin=-5,xmax=5,
tick label style={font=\footnotesize},
ytick style={color=black},
xtick = {-5,0,5},
ytick = {-5,0,5},
xticklabel={\pgfmathparse{\tick*100}\pgfmathprintnumber{\pgfmathresult}\%},
yticklabel={\pgfmathparse{\tick*100}\pgfmathprintnumber{\pgfmathresult}\%}
]
\draw (0.25,1.2) node[cross=4pt,rotate=45, semithick]{};

\addplot graphics [includegraphics cmd=\pgfimage,xmin=-10, xmax=10, ymin=-10, ymax=10] {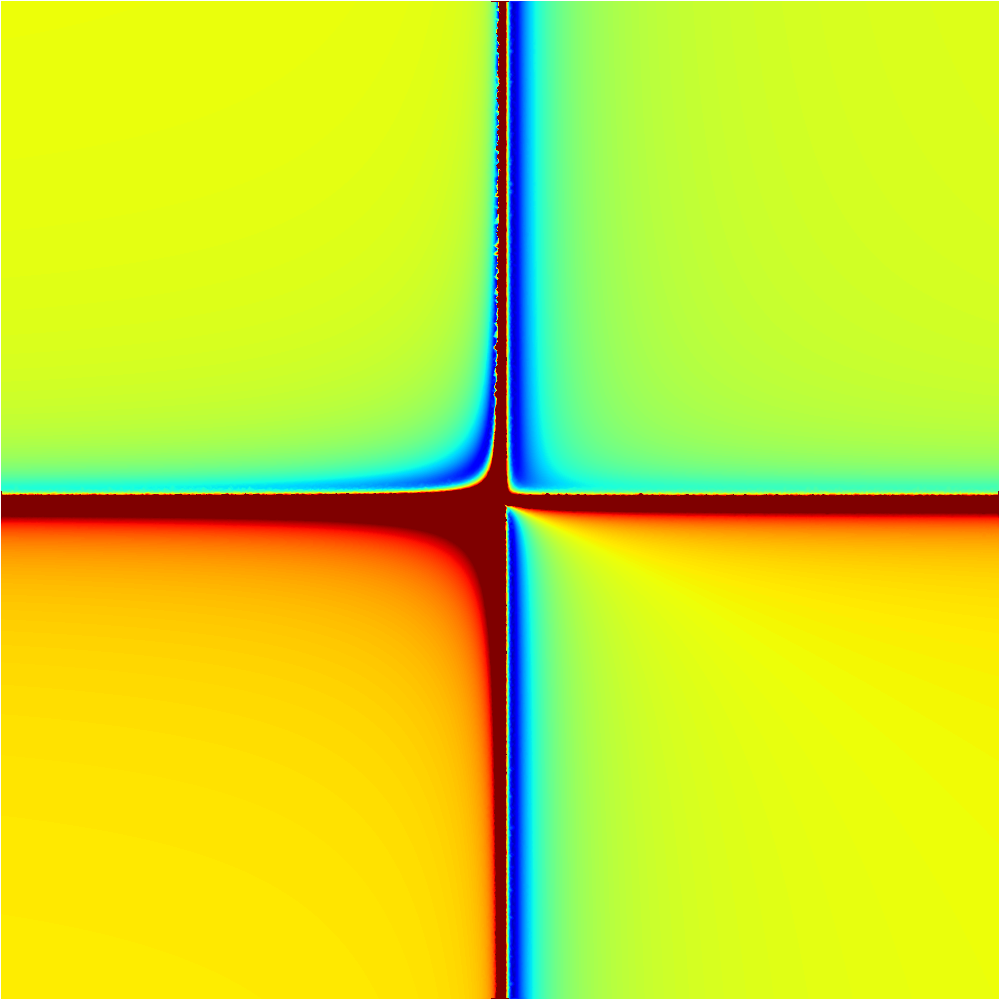};
\nextgroupplot[
tick align=outside, height = 6.5cm, width = 6.5cm, 
tick pos=left,
x grid style={white!69.0196078431373!black},
xtick style={color=black},
y grid style={white!69.0196078431373!black},
ymin=-3,ymax=2,xmin=-0,xmax=1,
xlabel={\footnotesize $\eta_s$},
xtick = {0,0.5,1},
tick label style={font=\footnotesize},
ytick style={color=black},
xticklabel={\pgfmathparse{\tick*100}\pgfmathprintnumber{\pgfmathresult}\%},
yticklabel={\pgfmathparse{\tick*100}\pgfmathprintnumber{\pgfmathresult}\%},
colorbar,
name = rightp,
colorbar style={
        title=\footnotesize RMSE,
        colormap name=jet,
        ytick={0,0.135,0.2789,0.442,0.72,1},
        yticklabels={\footnotesize $0.5$,\footnotesize $0.7$,\footnotesize $1.0$,\footnotesize $1.5$,\footnotesize $3.0$,\footnotesize $6.0$},
        scale = 1,
        at={(1.25,0.5)},
        anchor= west,
        colorbar/width=2.5mm,
        yticklabel style={
            text width=1.3em,
            align=right,
            /pgf/number format/.cd,
                fixed,
                fixed zerofill
        }}
]
\addplot graphics [includegraphics cmd=\pgfimage,xmin=-10, xmax=10, ymin=-10, ymax=10] {plots/marg/margFalseTrue-002.png};
\draw (0.25,1.2) node[cross=4pt,rotate=45, thick]{};
\draw (0.26,-2.29) node[cross=4pt,rotate=45, thick, white]{};
\end{groupplot}
\end{tikzpicture}
    \caption[\acrshort{rmse} heatmap for MARG68.]{Heatmap for MARG68 in [\si{\joule\per\kilo\gram\per\meter}] showing the relationship between the shortening $\eta_s$ and lengthening $\eta_l$ efficiency and the RMSE. On the right zoomed-in heatmap, the original MARG68 model is pointed out with a black cross, and the quasi-optimised MARG68 parameters are annotated with a white cross. }
    \label{fig:marg1}
\end{figure}
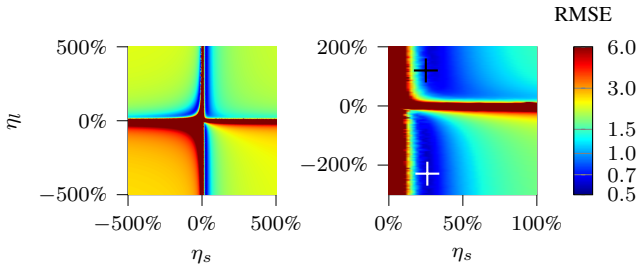

In MARG68, the KSstat sensitivity index $S_{i}$ for shortening is $0.50$ and higher than the one for lengthening with $0.30$. As MARG68 contains only two parameters, the RMSE for each parameter combination can be visualised in a 2D-plane (see Figure \ref{fig:marg1}). A minimum RMSE was found at a shortening efficiency $\eta_s$ of \SI{25.7}{\percent} and a lengthening efficiency $\eta_l$ of \SI{-228}{\percent}, corresponding to \SI{44}{\percent} energy restoration. 

\subsubsection*{MINE97}
\begin{figure}[h]
    \centering
    \input{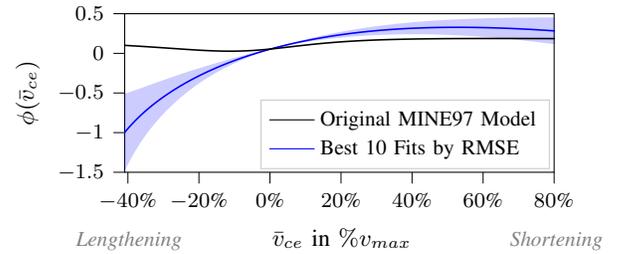}
    \caption[MINE97's $\phi$-function]
    {MINE97's $\phi$-function. Mean and standard deviation of the ten best fits from the \acrshort{mc} simulation and the original MINE97 model. Minimum and maximum values of $\Bar{v}_{ce}$ were chosen according to the values occurring in the dataset.}
    \label{fig:mimimine}
\end{figure}

Similar to MARG68, the MINE97 model also estimates the metabolic rate as a function of fibre velocity. In MINE97's polynomial $\phi$-function, the lower-order parameters ($p_{mi,1,4}$) generally have the highest sensitivity indices. Figure \ref{fig:mimimine} shows the best fitting $\phi$-functions from the MC simulation and compares it to the original MINE97 model. During shortening, the best fits estimate a higher metabolic rate, while during lengthening, the best fits predict negative metabolic rates.

\subsubsection*{LICH05}
As visible in Figure \ref{fig:glowchart}, LICH05 is the most complex MEE model in our study, containing a shortening-lengthening as well as a maintenance heat rate. The results show that the highest sensitivity index belongs to parameter $p_{l,8}$, associated with muscle shortening. The second highest sensitivity index belongs to parameter $p_{l,1}$, regulating the impact of Hill-type force-length relationship on the maintenance heat rate. On the other hand, the parameters of the $\gamma$-function of the LICH05 model, which are related to the activation time dependency, have the lowest sensitivity indices.

\subsubsection*{KIMR15}
In KIMR15, KSstats for maintenance heat rate $S_i = 0.91$ and co-contraction heat rate $S_i=0.83$ dominate. The sensitivity of the shortening-lengthening heat rate $S_i$ is $0.47$ for shortening and $0.31$ for lengthening.


\subsection{Deep MEE Models}\label{NNN}
We found that for muscle and joint space, the minimum achieved RMSEs were \SI{0.87}{\joule\per\kilo\gram\per\meter} and \SI{0.91}{\joule\per\kilo\gram\per\meter}, respectively. Therefore, none of the ANNs was able to achieve a better RMSE than the quasi-optimised MEE models.
In the muscle space, the feature combination with the lowest RMSE was contractile element velocity $\Tilde{v}_{ce}$, muscle stimulation $e$, activation $a$,  maximum isometric force $F_{max}$ and the fast-twitch fibre percentage ${r}_{ft}$.
The 20 best results used muscle activation, stimulation or both, and used fibre velocity $\Tilde{v}_{ce}$. Further features that were encountered among most of the top results were maximum isometric force $F_{max}$ and the fast twitch fibre percentage $r_{ft}$. 

The combination of joint space features with the lowest RMSE was the joint moment $M$, velocity $\dot{q}$ and acceleration $\ddot{q}$. Omitting $\ddot{q}$ increased the RMSE by \SI{7.2}{\percent}. Every other change in feature selection saw an RMSE increase of at least \SI{16.8}{\percent}. The RMSE achieved by the quasi-optimised MEE version of KIMR15 was \SI{26}{\percent} higher than the best-performing joint-space ANN.

\begin{figure*}[h!]
     \begin{minipage}[t]{0.48\linewidth}
         \centering
        \vspace{-8.1cm}
\begin{tikzpicture}

\begin{axis}[
tick align=outside,
title={(a) Joint Space},
tick pos=left,
x grid style={white!69.0196078431373!black},
xmin=-0.5, xmax=3.5,
xtick style={color=black},
xtick={0,1,2,3},
xticklabels={$\Dot{q}$,$q$,$M$,$\Ddot{q}$},
y grid style={white!69.0196078431373!black},
ymin=-0.5, ymax=3.5,
ytick style={color=black},
ytick={0,1,2,3},
yticklabels={$\Dot{q}$,$q$,$M$,$\Ddot{q}$},
]
\addplot graphics [includegraphics cmd=\pgfimage,xmin=-0.5, xmax=3.5, ymin=-0.5, ymax=3.5] {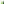};
\draw (axis cs:1,0) node[
  scale=1,
  text=black,
  rotate=0.0
]{1.26};
\draw (axis cs:2,0) node[
  scale=1,
  text=black,
  rotate=0.0
]{0.97};
\draw (axis cs:3,0) node[
  scale=1,
  text=black,
  rotate=0.0
]{1.67};
\draw (axis cs:2,1) node[
  scale=1,
  text=black,
  rotate=0.0
]{1.19};
\draw (axis cs:3,1) node[
  scale=1,
  text=black,
  rotate=0.0
]{1.76};
\draw (axis cs:3,2) node[
  scale=1,
  text=black,
  rotate=0.0
]{1.21};
\end{axis}

\end{tikzpicture}
     \end{minipage}
     \hspace{0.5cm}
     \centering
     \begin{minipage}[t]{0.48\linewidth}
         \hspace{-1cm}
         \centering
\begin{tikzpicture}

\begin{axis}[
tick align=outside,
tick pos=left,
x grid style={white!69.0196078431373!black},
xmin=-0.5, xmax=7.5,
xtick style={color=black},
xtick={0,1,2,3,4,5,6,7},
xticklabel style={rotate=90.0},
xticklabels={$\Tilde{l}_{ce}$,$\Tilde{v}_{ce}$,$a$,$F_{max,iso}$,$w_h$,$r_{ft}$,$e$,$l_{ce,opt}$},
y grid style={white!69.0196078431373!black},
ymin=-0.5, ymax=7.5,
ytick style={color=black},
title = {(b) Muscle Space},
ytick={0,1,2,3,4,5,6,7},
yticklabels={$\Tilde{l}_{ce}$,$\Tilde{v}_{ce}$,$a$,$F_{ iso}$,$w_h$,$r_{ft}$,$e$,$l_{ce,opt}$},
]
\addplot graphics [includegraphics cmd=\pgfimage,xmin=-0.5, xmax=7.5, ymin=-0.5, ymax=7.5] {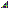};
\draw (axis cs:1,0) node[
  scale=1,
  text=white,
  rotate=0.0
]{2.06};
\draw (axis cs:2,0) node[
  scale=1,
  text=black,
  rotate=0.0
]{1.23};
\draw (axis cs:3,0) node[
  scale=1,
  text=white,
  rotate=0.0
]{1.80};
\draw (axis cs:4,0) node[
  scale=1,
  text=white,
  rotate=0.0
]{1.87};
\draw (axis cs:5,0) node[
  scale=1,
  text=white,
  rotate=0.0
]{1.97};
\draw (axis cs:6,0) node[
  scale=1,
  text=black,
  rotate=0.0
]{1.25};
\draw (axis cs:7,0) node[
  scale=1,
  text=white,
  rotate=0.0
]{2.01};
\draw (axis cs:2,1) node[
  scale=1,
  text=black,
  rotate=0.0
]{0.91};
\draw (axis cs:3,1) node[
  scale=1,
  text=white,
  rotate=0.0
]{1.99};
\draw (axis cs:4,1) node[
  scale=1,
  text=white,
  rotate=0.0
]{2.23};
\draw (axis cs:5,1) node[
  scale=1,
  text=white,
  rotate=0.0
]{2.03};
\draw (axis cs:6,1) node[
  scale=1,
  text=black,
  rotate=0.0
]{0.94};
\draw (axis cs:7,1) node[
  scale=1,
  text=white,
  rotate=0.0
]{1.96};
\draw (axis cs:3,2) node[
  scale=1,
  text=black,
  rotate=0.0
]{1.25};
\draw (axis cs:4,2) node[
  scale=1,
  text=black,
  rotate=0.0
]{1.32};
\draw (axis cs:5,2) node[
  scale=1,
  text=black,
  rotate=0.0
]{1.18};
\draw (axis cs:6,2) node[
  scale=1,
  text=black,
  rotate=0.0
]{1.23};
\draw (axis cs:7,2) node[
  scale=1,
  text=black,
  rotate=0.0
]{1.46};
\draw (axis cs:4,3) node[
  scale=1,
  text=white,
  rotate=0.0
]{2.01};
\draw (axis cs:5,3) node[
  scale=1,
  text=white,
  rotate=0.0
]{2.37};
\draw (axis cs:6,3) node[
  scale=1,
  text=black,
  rotate=0.0
]{1.36};
\draw (axis cs:7,3) node[
  scale=1,
  text=white,
  rotate=0.0
]{2.19};
\draw (axis cs:5,4) node[
  scale=1,
  text=white,
  rotate=0.0
]{2.21};
\draw (axis cs:6,4) node[
  scale=1,
  text=black,
  rotate=0.0
]{1.27};
\draw (axis cs:7,4) node[
  scale=1,
  text=white,
  rotate=0.0
]{2.48};
\draw (axis cs:6,5) node[
  scale=1,
  text=black,
  rotate=0.0
]{1.19};
\draw (axis cs:7,5) node[
  scale=1,
  text=white,
  rotate=0.0
]{2.21};
\draw (axis cs:7,6) node[
  scale=1,
  text=white,
  rotate=0.0
]{1.95};
\end{axis}

\end{tikzpicture}
     \end{minipage}
     \caption[Joint/Muscle Params]
     {
     {
     Comparison of RMSE [\si{\joule\per\kilo\gram\per\meter}] for joint (a) and muscle space (b) across various feature combinations. The RMSE corresponds to an ANN precisely trained only those two features. The color scale represents the error magnitude, with yellow indicating lower error and blue representing higher error.}}
     \label{fig:jfcs}
\end{figure*}

In summary, including power-related features played a crucial role in enabling ANNs to achieve accurate estimations of metabolic costs. However, the force-related features ($M$, $a$, $e$) alone already contribute to the moderate accuracies observed in Figure \ref{fig:jfcs}. In contrast, the velocity-related features ($\dot{q}$, $\Tilde{v}_{ce}$) show improved performance only when combined with force-related features. A full list of feature combinations and their achieved RMSEs can be found in the supporting document.

\section{Discussion}
This study examined the influence of MEE model parameters on the metabolic cost estimations and thus which muscle or joint states and parameters are crucial for the metabolic cost estimations of gait. 
The three muscle-space MEE models featured between two and ten empirical parameters. Table~\ref{tab:overallp} shows that the quasi-optimised version of the more complex LICH05 model (see Figure~\ref{fig:glowchart}) only slightly outperforms the quasi-optimised versions of the simpler models, similar to the results reported in~\cite{Koelewijn.2019}. We concluded that {for estimating metabolic costs in (healthy) walking, simpler metabolic models are sufficient}.
As the CMC of $0.92$ between the quasi-optimised MEE models indicates, the quasi-optimised MEE models converged towards a mutual MEE estimation for each muscle over time.
Figure \ref{fig:allplots} shows the curve shape similarity, especially between LICH05 and MINE97, suggesting that their formulation is well suited for fitting gait data.

{The local sensitivity analysis conducted in both muscle and joint space revealed that the parameters with the highest sensitivity were those related to work. Unsurprisingly, this finding was globally supported by the ANNs used in the study. Furthermore, force-related parameters proved valuable in traditional MEE models and ANNs, even without including velocity-related parameters. Therefore, the suggested separate maintenance heat rate by LICH05 and KIMR15 is valid. The four traditional MEE models might benefit from including fast-twitch fibre ratios, as this parameter was present in the highest-ranking ANN, suggesting its potential contribution. Fast-twitch fibre ratios have been employed in several other MEE models before \cite{Bhargava.2004,Houdijk.2006,Umberger.2003}.}


A key finding was that the MEE curves of the quasi-optimised MEE models tend to show higher peak positive metabolic rates. {In most cases, the original MEE models underestimated metabolic costs. Therefore, the higher metabolic rates are not surprising. Furthermore, the quasi-optimised MEE models also showed} negative metabolic rates during lengthening (see Fig.~\ref{fig:allplots}). As shortening and lengthening exist during the gait cycle, the timing of energy consumption is shifted from more extreme negative MEE rates towards peak positive MEE rates. As seen in Figures \ref{fig:marg1} and \ref{fig:mimimine}, the quasi-optimised version of the MEE models would have a negative efficiency, or $\phi$-function in MINE97's case, during lengthening.
As depicted in Figure 4, the models fail to estimate sufficient heat to compensate for the energy delivered to the muscle, {which is not thermodynamically reasonable.} True negative metabolic rates would refer to the synthesis of adenosine triphosphate, which is impossible~\cite{Miller.2014, Woledge.2003}. 
It is plausible that modeling errors have contributed to this behavior.
In particular, {muscle states were estimated via dynamic optimisation} using Hill's muscle model. {This muscle models} is based on measurements of isotonic contractions near resting length {and only accurate for such conditions}~\cite{Winter.2009}. {We found a possible explanation in the} winding filament theory~\cite{Nishikawa.2012}, which suggests better capabilities {for muscles to store elastic energy}, thus explaining the negative MEE rates we found, as energy {stored during lengthening and released during shortening.}
To eliminate the possib{i}lity of this observation being due to the dataset used, we repeated the Monte Carlo simulation on a dataset by Theunissen et al.~\cite{THEUNISSEN2022107915}. Here, we also found a negative lengthening efficiency for the quasi-optimised MEE models. 

In this work, {the ANNs could not outperform the quasi-optimised MEE models. We attribute the reason to the inexact supervision combined with the relatively small training data set. The utilization of inexact supervision was preferred to enable the ANNs to mimic the traditional MEE model, which estimates metabolic rate using muscle/joint states at each time point.
} {In the future, however,} ANNs could potentially outperform other MEE models.
For example, we trained a temporal convolutional network on marker trajectories and ground reaction forces that reached \SI{20}{\percent} {lower error} compared to any of the tested feature combinations, performing slightly worse than the quasi-optimised LICH05 model. Recurrent ANNs such as temporal convolutional networks, LSTMs or transformer ANNs can replicate time-dependent muscle properties, as attempted in the activation-time-dependency in BHAR04 and LICH05. Furthermore, using raw data such as marker trajectories and ground reaction forces {can prevent} error accumulation during processing. Potential error sources can be inconsistencies in inverse dynamics and improper personalisation of musculoskeletal models.

To get first insights into the robustness of the quasi-optimised MEE and ANN models, which were trained on the data of Koelewijn et al. \cite{Koelewijn.2018}, we tested the models with other publicly available datasets of Theunissen et al.~\cite{THEUNISSEN2022107915} and Pimentel et al.~\cite{Pimetel4}.
On both datasets, the quasi-optimised MEE models saw a slight in- or decrease in RMSE {over their original counterparts}. ANNs did not generally outperform the quasi-optimised MEE models on these datasets either. This was expected since our training dataset was small and not necessarily representative for the conditions measured in those studies. 
It is important to note that the inter-MEE model discrepancy is generally less pronounced than the inter-dataset discrepancy in terms of metabolic cost estimation accuracy, likely due to different processing methods and settings.


In addition to more accurate MEE models, further advancements in biomechanical processing are necessary to improve metabolic cost estimations. Reconstruction through inverse kinematics and inverse dynamics with general musculoskeletal models leads to dynamic inconsistencies, where residual forces remain at the pelvis \cite{10.1115/1.4029304}. Processing differences between different models are large \cite{OpenSimVersusHumanBodyModelAComparisonStudyfortheLowerLimbsDuringGait}. To reduce both, a unified pipeline providing accurate model personalisation would be helpful. For example, the current scaling method implemented in OpenSim \cite{Delp.2007} does not personalise any muscle-related parameters such as maximum isometric forces, moment arms and fibre type percentages. However, the actual values of these muscle-related parameters vary vastly among the population~\cite{Powers.2012,PMID:1938097}. Most MEE models base their estimations on muscle-related parameters, implying the need for personalisation of these factors as well.

{
Based on our outcomes, we have defined some recommendations about which MEE model to choose for specific use cases. For instance, simpler MEE models can be used to calculate the metabolic cost of walking, as we have shown that they are similarly accurate as more complex models. However, it should be further tested if they can be used as objectives for walking simulations since computer optimizations might find unrealistic solutions when models are too simple. Furthermore, in other applications, especially those with more isometric contractions, we expect that more complex models are required. For example, MARG68 assigns no metabolic costs to isometric contractions and would likely provide inaccurate results. Similarly, the parameters of MINE97 are based on walking experiments. Nonetheless, we argue that the future development of MEE models should aim for agnosticism regarding specific movement types, while also focusing on minimal computational complexity to allow its application across a broad spectrum of scenarios.
}
 
\section{Conclusion}
We addressed the question which parameters and muscle/joint states are crucial in metabolic cost estimations.
{In the local sensitivity analysis}, we found that power-related parameters play the biggest role in metabolic cost estimations. Increasing the complexity of MEE models and the number of inputs does not necessarily increase the model's estimation accuracy. 

When exploring deep-learning as an alternative to classical MEE models, we could also globally confirm the non-necessity of many variables by training ANNs, because including some of the features only delivered marginally better {or even worse} results. 
{Using ANNs, did not achieve better accuracy than the quasi-optimised MEE models.}
Because we found that the inter-MEE model discrepancy is generally less pronounced than the inter-dataset discrepancy in terms of metabolic cost estimation accuracy, we argue for unified processing methods as a basis for validation of future work in MEE modelling.



\section*{Acknowledgment}
We thank Dr Dario Zanca and Dr An Nguyen for their advice on the deep learning approach. We thank Dr Jason Franz and Dr Richard Pimentel from the University of North Carolina at Chapel Hill for sharing their data and the general discussion about metabolic energy consumption.

\bibliographystyle{IEEEtran}

\bibliography{sources}







\end{document}